\documentclass[conference,compsoc]{IEEEtran}
\ifCLASSOPTIONcompsoc
  \usepackage[nocompress]{cite}
\else
  \usepackage{cite}
\fi
\ifCLASSINFOpdf
  \usepackage[pdftex]{graphicx}
\else
\fi
\usepackage{amsmath, amssymb}
\usepackage{algorithmic}

\usepackage{array}

\ifCLASSOPTIONcompsoc
  \usepackage[caption=false,font=footnotesize,labelfont=sf,textfont=sf]{subfig}
\else
  \usepackage[caption=false,font=footnotesize]{subfig}
\fi
\usepackage{url}

\usepackage{tabularx}
\usepackage{caption}
\usepackage{multirow}
\usepackage{wrapfig}

\begin{document}
%
\title{Spatiotemporal Graph Convolutional Recurrent Neural Network Model\\ for Citywide Air Pollution Forecasting}

\author{\IEEEauthorblockN{Van-Duc Le}
\IEEEauthorblockA{School of Electrical and\\Computer Engineering\\
Seoul National University\\
Seoul, Korea 08826\\
Email: levanduc@snu.ac.kr}
\and
\IEEEauthorblockN{Tien-Cuong Bui}
\IEEEauthorblockA{School of Electrical and\\Computer Engineering\\
Seoul National University\\
Seoul, Korea 08826\\
Email: cuongbt91@snu.ac.kr}
\and
\IEEEauthorblockN{Sang-Kyun Cha}
\IEEEauthorblockA{Graduate School of Data Science\\
Seoul National University\\
Seoul, Korea 08826\\
Email: chask@snu.ac.kr}}


%


\maketitle

\begin{abstract}
Citywide Air Pollution Forecasting tries to precisely predict the air quality multiple hours ahead for the entire city. This topic is challenged since air pollution varies in a spatiotemporal manner and depends on many complicated factors. Our previous research \cite{BigComp} has solved the problem by considering the whole city as an image and leveraged a Convolutional Long Short-Term Memory (ConvLSTM) model to learn the spatiotemporal features. However, an image-based representation may not be ideal as air pollution and other impact factors have natural graph structures. In this research, we argue that a Graph Convolutional Network (GCN) can efficiently represent the spatial features of air quality readings in the whole city. Specially, we extend the ConvLSTM model to a Spatiotemporal Graph Convolutional Recurrent Neural Network (Spatiotemporal GCRNN) model by tightly integrating a GCN architecture into an RNN structure for efficient learning spatiotemporal characteristics of air quality values and their influential factors. Our extensive experiments prove the proposed model has a better performance compare to the state-of-the-art ConvLSTM model for air pollution predicting while the number of parameters is much smaller. Moreover, our approach is also superior to a hybrid GCN-based method in a real-world air pollution dataset.
\end{abstract}


%
\IEEEpeerreviewmaketitle

\section{Introduction}
\label{intro}
Air Pollution is a severe problem for many big cities in the world. Accurately predicting air quality multiple hours ahead is a challenging task in recent years. One of the most concerning problems is that air pollution varies by both spatial and temporal forms. As pointed in recent research papers \cite{YuZheng, BigComp, BIC2018, Alex2018Deep, Alex2020Star}, the air cleanness in a city changes from one location to other locations and time by time. Therefore, we need a \textit{spatiotemporal architecture} to model air pollution features efficiently and effectively.

Our previous paper in \cite{BigComp} illustrated that many \textit{spatiotemporal factors} can affect a city's air quality. These factors include meteorological factors such as rain, humidity, wind speed, wind direction; transportation factors like the traffic volume or vehicle driving speed; and external factors such as air pollution from nearby cities or areas. These mentioned spatiotemporal influences make the air pollution forecasting problem harder; thus, an air quality prediction model should effectively represent the air pollution values and their spatiotemporal impact factors. Our past research introduced a large-scale \textit{real-world air pollution dataset} collected from Seoul city in South Korea to tackle the spatiotemporal air pollution forecasting task. Using the collected Seoul data, we proposed to use a combination of Convolutional Neural Network (CNN) and Long Short-Term Memory (LSTM) in a \textit{ConvLSTM model} to interpolate and predict air pollution for the entire city by leveraging an image-based approach. The ConvLSTM model outperformed other Deep Learning models in forecasting air pollution of Seoul city in 12 hours ahead. However, the image-based representation may not capture well the natural graph structures of air pollution observation stations in a city and other impact factors such as meteorology or traffic. This research tries to solve and overcome the mentioned problem by leveraging a graph-based method.

\begin{figure}[htbp]
\centering
  \includegraphics[width=\linewidth]{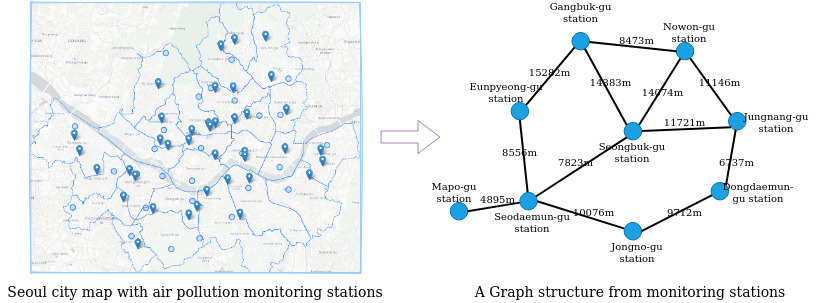}
\caption{A Graph structure constructed for the air pollution forecasting problem in Seoul city, South Korea. Each node denotes an air pollution monitoring station. The labels of edges are the real-world distances between stations (some are omitted).}
\label{fig:graph} 
\end{figure}

In Fig. \ref{fig:graph}, we show a graph structure constructed for the air pollution forecasting problem in Seoul city, South Korea. Each node denotes an air pollution monitoring station, and edges are the connections between two stations weighted by their distance. Inspired from recent success in using Graph Convolutional Networks (GCNs) for spatiotemporal problems and derived from the spatiotemporal basis of air pollution data and its influential factors, we propose a \textbf{Spatiotemporal Graph Convolutional Recurrent Neural Network (Spatiotemporal GCRNN)} model. Our model uses a GCN structure to encode the spatial feature and an RNN architecture to model the time-series features of the air pollution data. The most crucial point is that we tightly integrate the GCN architecture into the RNN structure to have a unified unit that efficiently learns the spatiotemporal features at the same time. Our experiments prove that the proposed method yields better performance than previous approaches to Seoul data while the parameters are much smaller.

In the ConvLSTM paper for air pollution interpolating and forecasting \cite{BigComp}, a large scale dataset of air pollution and spatiotemporal related factors was constructed. This dataset includes air pollution values in monitoring stations collected in Seoul city for three years, from 2015 to 2017, and many air quality impact factors that contribute to building an accurate air pollution prediction model. However, we believe that an even larger dataset will accelerate this field of research with better prediction models. Therefore, in this study, we collect more data from Seoul city in two recent years of 2018 and 2019 with all aforementioned spatiotemporal factors. The new dataset has more extended period of data (from 2015 to 2019), more data points (a 75$\%$ bigger) which brings a more significant training data for both air pollution and spatiotemporal research.

In summary, our contributions in this paper are three folds:
\begin{itemize}
    \item \textbf{Firstly}, we introduce \textbf{Spatiotemporal Graph Convolutional Recurrent Neural Network model}, a unified integration of GCN and RNN architectures, which works efficiently and effectively for spatiotemporal air pollution forecasting problems.
    \item \textbf{Secondly}, we conduct extensive experiments to prove the proposed method produces \textbf{better performance than the state-of-the-art ConvLSTM model} and \textbf{outperforms a recent GCN based approach}. We implemented the two most common graph convolutional operators, such as spectral graph and diffusion graph convolution, in our model.
    \item \textbf{Lastly}, we \textbf{enlarge the previous large-scale dataset of Seoul data to a new massive dataset} to bring researchers a large-scale data source for both air pollution and spatiotemporal related problems.
\end{itemize}

The rest of the paper is structured as follows: First, we review relating literature in the related work section. The following are detailed descriptions of our proposed method and architecture. The empirical experiments are then presented and evaluated. Finally, we conclude and discuss our future research directions.

\section{Related work}
\label{related}
\subsection{Spatiotemporal Deep Learning for Air Pollution Forecasting}
Many Deep Learning based models have been adopted for spatiotemporal air pollution prediction. In \cite{YuZheng}, the authors proposed a method that used a \textit{spatial predictor} as a neural network to model the spatial correlations at different locations. Their approach defined spatial neighbors of a station within three circles from farthest (e.g., 300km) to nearest (e.g., 30km), and then aggregated the air pollutant values and related factors such as meteorological data in these circles. Their spatial learning algorithm is complicated and uses hand-crafted features (e.g., the circles' diameters). In contrast, our approach uses graph convolution to automatically learn the graph structures of air pollution stations and related factors in any locations in a city.

Similarly, in \cite{BIC2018}, the authors introduced a \textit{real-time air pollution prediction model} based on big data; and in \cite{Alex2018Deep}, an \textit{LSTM-based encoder-decoder method} was employed to forecast air pollution in South Korea. A more recent research \cite{Alex2020Star} models air pollution influential factors using a \textit{multi-modal approach} for air pollution forecasting in Seoul city. Nevertheless, none of those aforementioned solutions leverages the graph structure for air pollution data as in our current research.

Air pollution forecasting is also a time series prediction problem. There are some well-known machine learning algorithms for time series forecasting such as Auto-regressive Integrated Moving Average (ARIMA), Support Vector Regression (SVR), and Recurrent Neural Networks (RNNs). In \cite{Alex2020Star}, the authors did experiments to show that the ARIMA, SVR, and pure RNN models were inefficient in middle-to-long-term air pollution prediction since they could not exploit spatiotemporal information from many impact factors. On the other hand, some RNN variants such as Long Short-Term Memory (LSTM) \cite{LSTM} and Gated Recurrent Units (GRUs) \cite{GRU} enable us to handle longer training sequence and create more accurate prediction for long-term foreseeing. Many recent papers have leveraged and proved the advantages of LSTM and GRUs structures in air pollution forecasting \cite{BigComp, Alex2018Deep, Alex2020Star}. In this paper, we follow previous papers and propose to use GRUs as an RNN architecture to learn temporal features of air pollution data.

\subsection{Spatiotemporal Graph Convolutional Networks}
In recent years, Convolutional Neural Networks (CNNs) have been generalized to arbitrary graphs based on the \textit{spectral graph theory}. Graph Convolutional Networks (GCNs) were first introduced in \cite{SpectralGraph}, which connects the spectral graph theory and deep neural networks. Paper \cite{ChebNet} proposes \textit{ChebNet} model to improve GCN with fast localized spectral convolutional filters. Paper \cite{SimplifiedChebNet} \textit{simplifies ChebNet} by parameterizing each spectral filter by the first order Chebyshev polynomials and achieves state-of-the-art performance in semi-supervised classification tasks.

GCNs are now trending for spatiotemporal problems like traffic forecasting. In \cite{DCRNN}, the authors represented the pair-wise spatial correlations between traffic sensors using a directed graph in which nodes are sensors, and edge weights are closeness between the sensor pairs denoted by the roads network distance. The proposed model was a \textit{graph diffusion convolution model} to capture the spatial dependency of traffic sensors. Many following papers also suggest using GCN in traffic forecasting and produce reasonable results \cite{STGCN, DySTGCNN}.

Similarly, GCNs have been started using in air pollution forecasting problems. A paper in \cite{LinY} proposed a \textit{geo-context based diffusion convolutional recurrent neural network (GC-DCRNN)} model for forecasting short-term PM$_{2.5}$ concentrations. In the paper \cite{Hybrid}, the researchers introduced to use a \textit{sequential combination of graph convolutional neural network and long short-term memory model} for spatiotemporal PM$_{2.5}$ forecasting in 72 hours ahead. A newer paper \cite{KnowledgeGNN} defined a \textit{knowledge enhanced graph neural network model} for PM$_{2.5}$ forecasting. These papers both leveraged the GCN model as the spatial representation for air pollution features but still have shortcomings compared to our approach.

The geo-context based method in \cite{LinY} tried to use geographic features around monitoring stations such as roads, buildings, green lands, or water areas to model the spatial relationships between stations. It also exploited a graph convolutional recurrent neural network as in our system. The biggest problem is that they used static data such as geographic features collected from OpenStreetMap data. In contrast, we use dynamic and real-time data like traffic volume and average vehicle speeds on roads. Moreover, in crowded cities like Seoul, the geo-context around each monitoring station may be similar because of the high density of buildings and roads. Therefore, the geo-context is reduced to the real-world distances between stations as in our method.

The approach in \cite{Hybrid} firstly extracted the spatial features of input data by graph convolution operations, and then fed the extracted features into an LSTM model to learn the temporal features. The final layer was a fully connected layer to regress the PM$_{2.5}$ values. This paper's input data was air pollution values and weather data in Jing-Jin-Ji (Beijing, Tianjin, and Hebei) area in North China. The distinct steps of GCN and LSTM layers in this method may not accurately capture the complex correlations of many spatiotemporal factors in the air pollution forecasting problem as in our Spatiotemporal GCRNN model can capture both spatial and temporal features at the same time. The more recent paper \cite{KnowledgeGNN} attempted to enhance the nodes and edges of a graph by the knowledge from the air pollution domain such as the air pollution transportation between cities, the meteorological information, and the geographical knowledge (e.g., mountains). We argue that their knowledge enhance approach does not generalize well for other spatiotemporal forecasting problems and they only consider the city-level prediction, whereas we try to forecast air pollution at the station-level for a city.

\section{Methodology}
\label{method}
In this section, we formalize the air pollution forecasting as a GCN problem and describe how the \textit{Spatiotemporal Graph Convolutional Recurrent Neural Network} model can capture well the spatiotemporal features.

\subsection{Graph and Graph Convolutional Network Representations}
The most important concepts in the graph theory are (Weighted) Adjacency Matrix ($W$), Degree Matrix ($D$), and Laplacian Matrix ($L$). For a graph $G=(V,E)$, in which $V=\{v_i\}$, $i=1,..,N$, $E=\{(v_i,v_j)\}$, the following equations define $W$, $D$, $L$ of a graph, and some common types of Laplacian Matrices such as symmetric normalized Laplacian (L$^{sym}$) or random walk normalized Laplacian (L$^{rw}$):
\begin{align}
\label{eqn:graph_definition}
    &W = \{w_{ij}\}, w_{ij} = \begin{cases} \text{weighted value, if } (v_i,v_j) \in E \\
    0 \text{, otherwise} \end{cases} \nonumber \\
    &D = \{d_{ij}\}, d_{ij} = \begin{cases} deg(v_i) \text{, if i = j}, \sum deg(v_i) = 2|E| \nonumber \\
    0 \text{, otherwise} \end{cases} \nonumber \\
    &L = D - W  \\
    &L^{sym} = D^{-1/2}(D-W)D^{-1/2} \nonumber \\
    &L^{rw} = D^{-1}(D-W) \nonumber
\end{align}

The following equation denotes the kernel of a GCN in the ChebNet model \cite{ChebNet}:
\begin{equation}
\label{eqn:kernel_chebnet}
    g_\theta(\Lambda) = \sum_{k=0}^{K-1} \theta_k T_k (\tilde{\Lambda})
\end{equation}
where $g_\theta$ is a kernel ($\theta$ represents the vector of Chebyshev coefficients) applied to $\Lambda$, the diagonal matrix of Laplacian eigenvalues ($\tilde{\Lambda}$ represents the diagonal matrix of scaled Laplacian eigenvalues, $\tilde{\Lambda} = 2\Lambda/\lambda_{max} -I_n$, $\lambda_{max}$ is the largest eigenvalue of $L$). $k$ is the smallest order neighborhood, and $K$ indicates the largest order localization. Finally, $T$ stands for the Chebyshev polynomials of the $k$th order.

\subsection{Problem Formalization}
\label{problem}
\begin{figure}
\centering
  \includegraphics[width=2.5in]{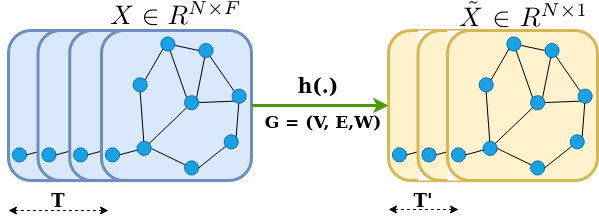}
\caption{An air pollution forecasting task is constructed as a graph-based problem. $T$ is the number of historical graph signals, $T'$ is the number of future prediction graph signals. $X \in R^{N \times F}$ is an input graph signal, $\tilde{X} \in R^{N \times 1}$ is an output signal, $h(.)$ is the learned function given the graph $G$.}
\label{fig:graph_problem} 
\end{figure}

The goal of an air pollution forecasting task is to predict the future air quality values given previously observed air pollution from $N$ correlated air pollution monitoring stations. We can represent the monitoring stations network as a weighted graph $G = (V, E, W)$, where $V$ is a set of nodes or stations, $|V| = N$, $E$ is a set of station relationships (edges), and $W \in R^{N \times N}$ is a weighted adjacency matrix denoting the stations correlations. The distance of two observation stations indicates the relation between two stations in the graph. Let $F$ is the number of features in each node (e.g., the air pollution values and other impact factors values at that node), then the input graph signal of $G$ is $X \in R^{N \times F}$. $X^{(t)}$ denotes the input graph signal observed at time $t$, and $\tilde{X}^{(t')}$ is the output graph signal at time $t'$, $\tilde{X} \in R^{N \times 1}$ (because we only output one estimated air pollution value for one station at a time). The air pollution forecasting problem tries to learn a function $h(.)$ that maps $T$ historical graph signals to future $T'$ graph signals, given a graph $G$:
\begin{equation}
\label{eqn:problem}
    [X^{(t-T+1)}, ..., X^{(t)}; G] \xrightarrow{h(.)} [\tilde{X}^{(t+1)}, ..., \tilde{X}^{(t+T')}]
\end{equation}
Fig. \ref{fig:graph_problem} shows how an air pollution forecasting task is constructed as a graph based problem with the graph signals representing the air pollution monitoring stations. Both input and output graph signals have the same graph structure.

\subsection{Modeling Spatial Features}
\label{spatial_modeling}
In this paper, we learn the \textit{spatial dependency} between air pollution monitoring stations by representing the station network into a graph structure as in the Problem Formalization section. After that, we employ the graph convolution operators to model the spatial features of input graph signals. The extracted spatial features are used in combination with temporal features to predict future air pollution outputs. We try with two well-known graph convolution definitions, spectral graph convolution and diffusion graph convolution, for modeling spatial dependency. The formulas of two graph convolutional definition are presented below.

The \textit{spectral graph convolution} \cite{SpectralGraph, ChebNet, SimplifiedChebNet} employs the concepts of symmetric normalized graph Laplacian matrix $L = D^{-1/2}(D-W)D^{-1/2} = \Phi\Lambda\Phi^T$. The convolution operation over a graph signal $X \in R^{N \times F}$ and a kernel $g_\theta$ is defined as below:
\begin{align}
\label{eqn:chebnet}
    X_{:,f} \star_G g_\theta = \Phi(\sum_{k=0}^{K-1} \theta_k \Lambda^k)\Phi^T X_{:,f} \nonumber \\
    = \sum_{k=0}^{K-1} \theta_k L^k X_{:,f} = \sum_{k=0}^{K-1} \tilde{\theta_k}T_k(\tilde{L})X_{:,f}
\end{align}
with the kernel $g_\theta$ as in equation \ref{eqn:kernel_chebnet}, $\theta$ are the learnable parameters, $f \in \{1, ..., F\}$. $\star_G$ denotes the convolution over a graph G. $T_0(x) = 1, T_1(x) = x, T_k(x) = xT_{k-1}(x) - T_{k-2}(x)$ are the basis of the Chebyshev polynomials. $\tilde{L} = 2L/\lambda_{max} -I$ represents a rescaling of the graph Laplacian that maps the eigenvalues from $[0, \lambda_{max}]$ to $[-1, 1]$ since Chebyshev polynomial forms an orthogonal basis in $[-1, 1]$. We can approximate the kernel $g_\theta$ by a truncated expansion in terms of Chebyshev polynomials $T_k(x)$ up to $K$th order. The graph convolution definition is now have the form:
\begin{equation}
\label{eqn:spectral_conv}
    X_{:,f} \star_G g_{\theta'} \approx \sum_{k=0}^{K} \theta'_k T_k(\tilde{L})X_{:,f}
\end{equation}
This expression is $K$-localized since it is a $K$th-order polynomial in the Laplacian, i.e. it depends only on nodes that are at maximum $K$ steps away from the central node ($K$th-order neighborhoods). Equation \ref{eqn:spectral_conv} is our spectral graph convolution definition used in this paper.

A \textit{graph diffusion convolution} was defined in the paper \cite{DCRNN}. The diffusion process is characterized by a random walk on graph $G$ with restart probability $\alpha \in [0, 1]$, and a state matrix $D^{-1}W$, where $D$ is the degree diagonal matrix (see equation \ref{eqn:graph_definition}). The resulted diffusion convolution from the mentioned diffusion process is specified as:
\begin{equation}
\label{eqn:diff_conv}
    X_{:,f} \star_G g_{\theta} = \sum_{k=0}^{K-1} (\theta_k (D^{-1}W)^k) X_{:,f} \text{ for $f \in \{1, ..., F\}$}
\end{equation}
where $\theta$ are the parameters for the filter $g_\theta$ and $D^{-1}W$ represents the transition matrix of the diffusion process. The parameter $K$ is the number of diffusion steps from a node to its neighborhood, which corresponds to the $K$th-order in the spectral graph convolution.

Since spectral graph convolution and graph diffusion convolution are two common graph convolution operators, in this paper, we implement each operator for the GCN layer and do experiments to compare their influences to the prediction results.

\begin{figure}[htbp]
\centering
  \includegraphics[width=\linewidth]{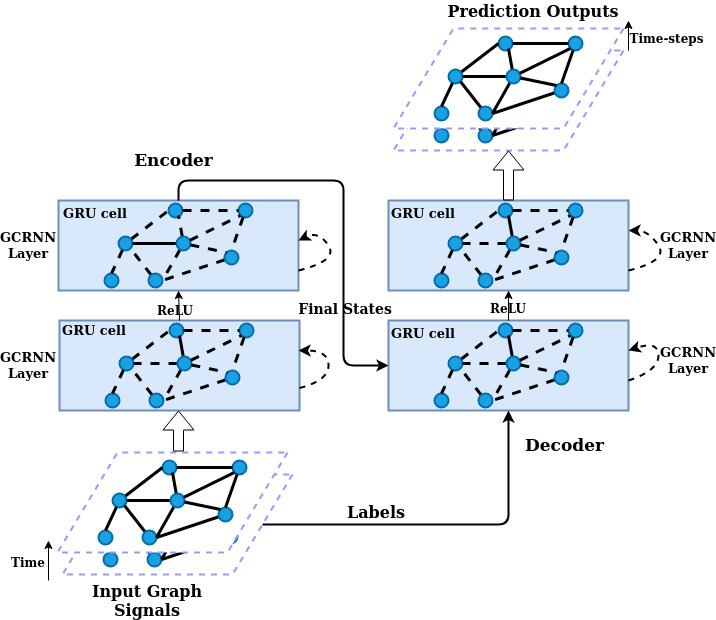}
\caption{The overall architecture of our \textbf{Spatiotemporal Graph Convolutional Recurrent Neural Network} model. Input Graph Signals are historical air pollution data represented in graph structures. Prediction Outputs are predicted air pollution values in some hours ahead. A GCRNN Layer includes a GRU cell with graph convolution operators. An Encoder and Decoder architecture is applied to predict multiple future time-steps.}
\label{fig:st_gcrnn} 
\end{figure}

\subsection{Modeling Temporal Features}
\label{temporal_modeling}
The previous paper \cite{BigComp} proved the power of a combination of CNN and RNN models in spatiotemporal forecasting problems by using convolutional operators to replace the fully connections in input-to-state and state-to-state transitions of an LSTM cell. In this paper, we also leverage that idea by replacing the matrix multiplications in a GRU cell with the graph convolution to \textit{learn spatial and temporal features of the input data simultaneously}. A GRU or Gated Recurrent Unit \cite{GRU} model is a variant of RNN model that can learn long-term sequence of data similar to an LSTM model but with the smaller number of parameters. A GRU cell with the graph convolution operators establishes a spatiotemporal layer that we call a \textbf{Graph Convolutional Recurrent Neural Network layer}, or \textbf{GCRNN layer}.

In detail, following are four transformation equations for reset gate, update gate, candidate values and hidden states in a GRU cell of one GCRNN layer.
\begin{align}
\label{eqn:gru}
    &r^{(t)} = \sigma(\Theta_r \star_G [X^{(t)}, H^{(t-1)}] + b_r) \nonumber \\
    &u^{(t)} = \sigma(\Theta_u \star_G [X^{(t)}, H^{(t-1)}] + b_u) \\
    &C^{(t)} = tanh(\Theta_C \star_G [X^{(t)}, (r^{(t)} \odot H^{(t-1)})] + b_C) \nonumber \\
    &H^{(t)} = u^{(t)} \odot H^{(t-1)} + (1 - u^{(t)}) \odot C^{(t)} \nonumber
\end{align}
where $X^{(t)}$, $H^{(t)}$ denote the input and output at time $t$, $r^{(t)}, u^{(t)}$ are reset gate and update gate at time $t$, $C^{(t)}$ are candidate values at time $t$, respectively. $\star_G$ means the graph convolutional operator defined by spectral graph convolution (equation \ref{eqn:spectral_conv}) or diffusion convolution (equation \ref{eqn:diff_conv}). $\Theta_r, \Theta_u, \Theta_C$ are parameters for the corresponding filters. $\odot$ denotes the Hadarmad product or element-wise matrix-matrix multiplication. $\sigma$ and $tanh$ are the activation functions.

For multiple steps forecasting, a sequence to sequence architecture (seq2seq) is applied \cite{Seq2Seq}. Both the encoder and decoder of the seq2seq consist of one or many GCRNN layers. The final states of the encoder are used to initialize the initial state of the decoder. The combination of GCRNN layers and a seq2seq scheme builds up our \textbf{Spatiotemporal Graph Convolutional Recurrent Neural Network (Spatiotemporal GCRNN)} model. Fig. \ref{fig:st_gcrnn} shows the system architecture of our Spatiotemporal GCRNN model designed for spatiotemporal air pollution forecasting. The input is the historical time-series of air pollution represented as graph signals, and the output is the future multiple hours of air pollution prediction.

\subsection{Modeling Spatiotemporal Impact Factors}
\label{spatiotemporal_factor}
As presented in previous paper \cite{BigComp}, air pollution in a city is influenced by many spatiotemporal factors. Regarding this paper's graph based approach, each spatiotemporal impact factor will be represented as a graph signal with the same graph structure as air pollution graph. We propose a strategy to learn these impact factors and air pollution graph features efficiently.

We fuse air pollution values and all spatiotemporal influential factors as one input graph signal with multiple features at the corresponding node. Denotes $X_a \in R^{N \times F_a}$ is the air pollution graph signal, $X_m \in R^{N \times F_m}$ is the meteorological graph signal, $X_t \in R^{N \times F_t}$ is the traffic volume graph signal, $X_s \in R^{N \times F_s}$ is the average driving speed graph signal, and $X_o \in R^{N \times F_o}$ is the outside air pollution graph signal. $N$ is the number of graph nodes, $F_a$, $F_m$, $F_t$, $F_s$, and $F_o$ are air pollution, meteorological, traffic volume, average speed, and outside air pollution number of features, respectively. Then the input graph signal $X$ of the model is a \textbf{concatenation} of all graph signals: 
\begin{equation}
\label{eqn:st_factors}
    X = X_a \oplus X_m \oplus X_t \oplus X_s \oplus X_o
\end{equation}
$\oplus$ is a vector concatenation operator. Therefore, the total number of features for the input graph signals is $\mathcal{F} = F_a + F_m + F_t + F_s + F_o$. All graph convolution operations (equation \ref{eqn:spectral_conv} and equation \ref{eqn:diff_conv}) and GRU equations (equation \ref{eqn:gru}) are not changed except the input graph signals turn to $X^{(t)} \in R^{N \times \mathcal{F}}$.

\begin{figure}[htbp]
\centering
  \includegraphics[width=\linewidth]{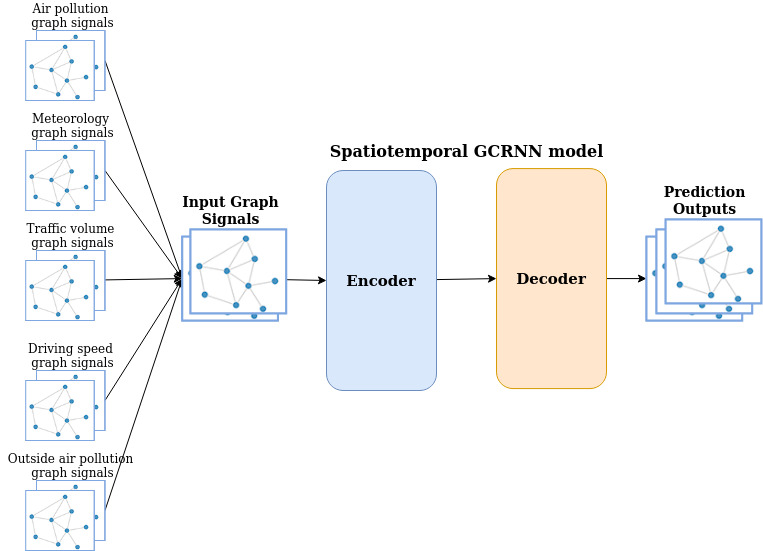}
\caption{Modeling spatiotemporal impact factors by fusing all input data into a combined graph signal.}
\label{fig:impact_factor}       
\end{figure}

Fig. \ref{fig:impact_factor} shows the strategy of modeling spatiotemporal impact factors as the combination of input graph signals. The architecture of the Spatiotemporal GCRNN model does not change.

\section{Experiments}
\label{experiments}
In this section, we describe our extensive experiments to prove the strength of the proposed Spatiotemporal GCRNN model in spatiotemporal air pollution forecasting tasks. After introducing the experiments' setup, we conduct experiments to directly compare the performance of the Spatiotemporal GCRNN to ConvLSTM, a state-of-the-art air pollution prediction method from the previous paper \cite{BigComp}, following are comparisons with a recent GCN-based air pollution forecasting model. Lastly, we ablation study the effects of many GCN configurations such as $K$th-order, weighted adjacency matrix construction, and different types of graph convolutional operators.

\subsection{Experiments Setup}
\label{experiment_setup}
We use the Seoul dataset as in the state-of-the-art paper \cite{BigComp}. The Seoul dataset is a large-scale dataset that consists of 3-year spatiotemporal data in Seoul city, Korea, from 2015 to 2017. This dataset includes air pollutants, such as $PM_{10}$, $PM_{2.5}$,...; meteorological data, like temperature, wind speed, wind direction, rainfall,...; traffic volume of main roads; average driving speed on roads; and the air pollution from 3 areas in China (Beijing, Shanghai, and Shandong) that affects Seoul's air quality. As mentioned in the Introduction section, in this paper, we broaden the Seoul dataset with a more extended period of air pollution and related impact factors data. Table \ref{tab:seoul_data} shows statistics about our new Seoul dataset compared to the existing one. Our new dataset is \textit{83\%} more in the number of air pollution samples and \textit{75\%} larger in the total number of all data points. Nevertheless, to keep a fair comparison with previous methods, we still do experiments with the existing Seoul dataset and reserve the new dataset for further research.

\begin{table}[htbp]
\renewcommand{\arraystretch}{1.2}
\caption{New Seoul dataset statistics and compared to the existing one}
\label{tab:seoul_data}
\begin{center}
\begin{tabularx}{\linewidth}{| p{4.3cm} | p{1.5cm} | p{1.5cm} |}
\hline Statistic & \textit{New dataset} & Existing dataset  \\
\hline Time Period  & \textit{01/01/2015 $\sim$ 12/31/2019} & 01/01/2015 $\sim$ 09/30/2017 \\
\hline \textbf{Total Hours} & \textit{43,824} & 24,048 \\
\hline \textbf{Total Number of Data Samples} & \textit{193,217,386} & 110,468,780 \\
\hline \textbf{Air Pollution Data Samples} & \textit{1,716,952} & 937,872 \\
\hline Meteorology Data Samples & \textit{753,326} & 735,734 \\
\hline Traffic Volume Data Samples & \textit{10,716,240} & 5,955,072 \\
\hline Average Driving Speed Data Samples & \textit{179,951,956} & 102,761,187 \\
\hline China Air Pollution Data Samples & \textit{78,912} & 78,912 \\
\hline
\end{tabularx}
\end{center}
\end{table}

We follow the same data pre-processing of ConvLSTM model from \cite{BigComp} for this paper's experiments, including the grid construction. Firstly, Seoul city is divided into a grid of shape $32 \times 32$, and each grid-cell is assigned with the corresponding observation stations or traffic survey points. This step makes our experiments in this paper comparable with the previous method. After the pre-processing, we construct a graph for the grid-level air pollution forecasting problem with each node is assigned by the corresponding cell.

We use the Tensorflow framework \cite{Tensorflow} to build the model and Nvidia GPU for training and testing. All experiments in this paper use the base learning rate of 0.001 and decay every ten epochs with the decay ratio 0.1 until reaching the minimal value of 2.0e-06. The batch size is 64, the number of GRU hidden units is 64, and the number of GCRNN layers is 2. We train with a maximum of 100 epochs but apply early stopping when the validation error does not decrease after 50 epochs. The Adam optimizer is utilized by its convergence speed and popular in deep learning optimizing problems. We also employ the \textit{Root Mean Squared Error (RMSE) loss} to compare the differences between the real observed and the estimated values. The training set is two years, 2015 and 2016, and the test set is the year 2017. This split strategy ensures the training and test set have the same distribution but still makes our model a good generalization.

Regarding test set errors, we apply the following metrics to have numerous viewpoints on the testing performance. The first metric is a common regression metric such as \textit{Root Mean Squared Error (RMSE)}. Also, we utilized a correlation coefficient \textit{R squared} ($R^2$) score, which was used in \cite{Hybrid}, to show how good is our prediction values' distribution fit to the real observed values. Moreover, the citywide spatiotemporal air pollution interpolation and prediction paper \cite{BigComp} presented a novel spatiotemporal metric called \textit{spRMSE} (or spatiotemporal RMSE). This metric evaluates how other spatiotemporal factors impact the efficiency of prediction outputs. To calculate this measurement, we use the following process: alternately remove one of the existing air pollution values in a node of the input graph signal in test set but still keep the features of other nodes and then compute RMSE of predicted air pollution value with the ground-truth one. The final error is the mean of RMSEs after this procedure with all nodes in the testing data. We will compare the \textit{spRMSE} metric values of our proposed Spatiotemporal GCRNN model to the ConvLSTM model.

\subsection{Performance comparisons with the ConvLSTM model for short-term air pollution forecasting}
\label{conv_lstm}

\begin{figure}[htbp]
\includegraphics[width=\linewidth]{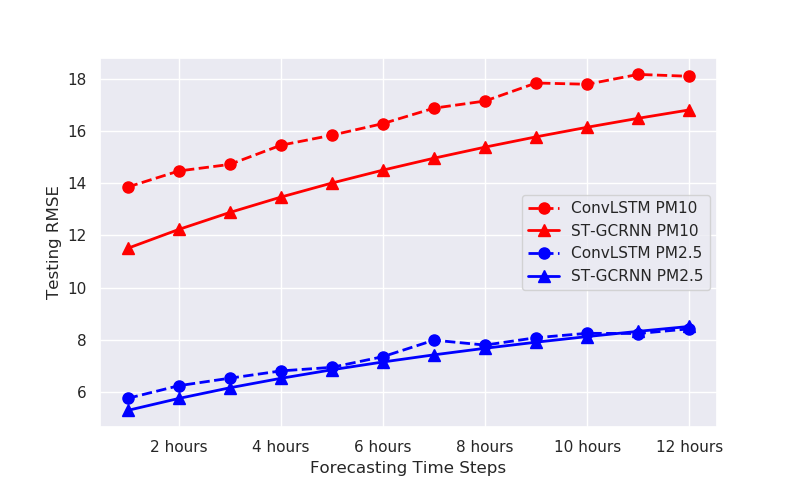}
\caption{Compare testing performance of ConvLSTM model (\textbf{ConvLSTM}) and Spatiotemporal GCRNN model (\textbf{ST-GCRNN}) for PM$_{10}$ and PM$_{2.5}$ air pollution forecasting from 1 to 12 hours in Seoul data (smaller is better).}
\label{fig:compare_convlstm}
\end{figure}

In this section, we conduct experiments of the Spatiotemporal GCRNN model for air pollution forecasting. We use PM$_{2.5}$ air pollutants as in the experiments of the ConvLSTM model \cite{BigComp}. The number of nodes of the input graph signals is 25 nodes (corresponding to 25 PM$_{2.5}$ monitoring stations). To prove the graph based model is better for spatial data, we also perform experiments with PM$_{10}$ air pollutants, which are observed by larger monitoring stations, 37 stations corresponding to 37 nodes of the input graph signals. The forecasting time-steps are similar to the paper \cite{BigComp}, from 1 hour to 12 hours in the future.

Fig. \ref{fig:compare_convlstm} shows RMSE metric values on testing data of the ConvLSTM model and Spatiotemporal GCRNN model with PM$_{2.5}$ and PM$_{10}$ air pollution forecasting from 1 hour to 12 hours (smaller is better). The Spatiotemporal GCRNN model produces \textbf{slightly better performance in RMSE values} compare to the ConvLSTM model with PM$_{2.5}$ air pollution prediction but achieves \textbf{significant better results for PM$_{10}$ prediction}. These results claim that a graph-based method can capture more information from a larger and more complex structure than a image-based approach in the ConvLSTM paper \cite{BigComp}.
Moreover, one advantage of the graph-based model is the smaller model size. The ConvLSTM model with 12 hours forecasting has the number of trainable parameters more than 20.5M parameters while the Spatiotemporal GCRNN model has only 371K parameters, a \textbf{55x smaller}. Hence, we can conclude that \textit{the Spatiotemporal GCRNN model with a much smaller size than a ConvLSTM model can produce better performance for spatiotemporal air pollution forecasting}.

Regarding comparing the performance of modeling the \textit{spatiotemporal impact factors} by ConvLSTM and Spatiotemporal GCRNN models, we use both \textit{RMSE} and \textit{spRMSE} metrics. We perform the experiments with the two types of input graph signals, air pollution only and air pollution with spatiotemporal impact factors, respectively. The performance results of Spatiotemporal GCRNN and ConvLSTM models are presented in Table \ref{tab:impact_factors}. In the table, \textit{ConvLSTM} and \textit{ConvLSTM + All} are ConvLSTM models without and with spatiotemporal impact factors data, \textit{ST-GCRNN} and \textit{ST-GCRNN + All} are Spatiotemporal GCRNN models including only air pollution input data and including all spatiotemporal impact factors, respectively. All experiments are executed with PM$_{2.5}$ air pollution 1-hour ahead prediction (similar to experiments in \cite{BigComp}). The results reveal that a \textit{Spatiotemporal GCRNN} model obtains better performance in terms of RMSE and spRMSE metrics than a ConvLSTM model in both input data are air pollution only and air pollution with all spatiotemporal impact factors. Even compare to the best RMSE in \cite{BigComp} with ConvLSTM + Meteorology data, the ST-GCRNN model still achieves smaller RMSE value (5.5947 vs. 6.5809). In addition, the \textit{Spatiotemporal GCRNN + All} has a \textit{smaller spRMSE} value than Spatiotemporal GCRNN without spatiotemporal impact factors. Therefore, we still recognize the efficacy of a GCN-based model in learning spatiotemporal factors for air pollution forecasting problems.

\begin{table}
\renewcommand{\arraystretch}{1.2}
\caption{Performance comparisons with spatiotemporal (ST) impact factors by ConvLSTM and Spatiotemporal GCRNN (ST-GCRNN) models on PM$_{2.5}$ with 2 metrics, spRMSE and RMSE (smaller is better).}
\label{tab:impact_factors} 
\centering
\begin{tabularx}{\linewidth}{| p{1cm} | p{1.2cm} p{1.2cm}| p{1.5cm} p{1.5cm} |}
\hline & \multicolumn{2}{c|}{Air pollution only} & \multicolumn{2}{c|}{With ST impact factors} \\
\hline Metric & ConvLSTM & ST-GCRNN & ConvLSTM + All & ST-GCRNN + All  \\
\hline spRMSE & 15.4872 & \textbf{14.9522} & 11.0254 & \textbf{10.9960} \\
\hline RMSE   & 8.3147 &  \textbf{5.5947}  & 7.1703  & \textbf{5.5963} \\
\hline
\end{tabularx}
\end{table}

\subsection{Performance comparisons with a GCN-based air pollution prediction model for long-term forecasting}
\label{hybrid_model}
In this experiment section, we choose the model in \cite{Hybrid} as our baseline for GCN-based air pollution forecasting comparisons. As pointed out in the related work section, the GCN-based model in \cite{Hybrid} is a Hybrid model that includes a GCN model and an LSTM layer as our system but it separates them to two consequential steps while we combine two architectures into a unified system.

We implement the \textit{Hybrid model of GCN and LSTM} from the paper \cite{Hybrid} in Seoul data. This model includes three sequential layers, the first layer is a GCN model to extract spatial features of the input data, the second one is an LSTM model with input is the extracted spatial features and output is the learned temporal features, and the last layer is a fully connected (FC) layer to regress the predictions. We implement GCN and LSTM layers of the Hybrid model with the same hyper-parameters as in our Spatiotemporal GCRNN model. We also apply the same training and testing configurations as in section \ref{experiment_setup}. The input graph signals of the Hybrid model are identical to our model.

For performance comparisons of this Hybrid model with our proposal model, we conduct experiments for medium to long-term prediction time steps: forecasting 12, 24, and up to 72 hours in the future with both PM$_{2.5}$ and PM$_{10}$ air pollutants. Table \ref{tab:hybrid_comparison} demonstrates the results on the same test set of \textit{the Hybrid model} and \textit{Spatiotemporal GCRNN} with PM$_{2.5}$ and PM$_{10}$ air pollution forecasting in 12, 24 and 72 hours ahead. Overall, our Spatiotemporal GCRNN model achieves better performance in PM$_{2.5}$ and PM$_{10}$ forecasting over the Hybrid model. In detail, our ST-GCRNN model obtains a larger gap of the performance in PM$_{2.5}$ forecasting than PM$_{10}$. The reason could be that the distributions of PM$_{2.5}$ air pollution values are more complicated than the PM$_{10}$ readings and our unified combination of GCN and GRU models can exploit the complex correlations of PM$_{2.5}$ air pollution better than sequential steps as in the Hybrid model. 

\begin{table}
\renewcommand{\arraystretch}{1.2}
\caption{Performance comparisons between the Hybrid model, and Spatiotemporal GCRNN (ST-GCRNN) in PM$_{2.5}$ and PM$_{10}$ for 12, 24, and 72 hours air pollution forecasting. The metric is RMSE (smaller is better).}
\label{tab:hybrid_comparison} 
\begin{tabular}{| l|c|p{2.5cm}|p{2.5cm} |}
\hline & T & Hybrid model & \textit{ST-GCRNN} \\
\hline \multirow{3}{3em}{PM$_{2.5}$} & 12 hours & 9.1880 & \textbf{8.5074} \\ 
\cline{2-4} & 24 hours & 12.9689 & \textbf{10.1350} \\
\cline{2-4} & 72 hours & 14.1043 & \textbf{12.5671} \\
\hline \multirow{3}{3em}{PM$_{10}$} & 12 hours & 16.8503 & \textbf{16.8093} \\
\cline{2-4} & 24 hours & 19.7901 & \textbf{19.1588} \\
\cline{2-4} & 72 hours & 24.4862 & \textbf{24.4360} \\
\hline
\end{tabular}
\end{table}

We prove the above assessment by doing statistical analysis for PM$_{2.5}$ and PM$_{10}$ air pollutants in Table \ref{tab:stat_analysis}. We compute the mean and standard deviation values for the following statistical measurements of the data: data points changing by time (which means the distribution of air pollution readings at each station over the time), and data points changing by locality (equals to the distribution of air pollution values over all stations in a city at a time). The statistic values are computed based on the [0-1] normalized data of PM$_{2.5}$ and PM$_{10}$ readings. From the table, the PM$_{2.5}$ air pollutant data has a higher standard deviation than the PM$_{10}$ regarding changing by time or by locality. It means the distribution of PM$_{2.5}$ values spreads around the mean value more than PM$_{10}$ data; thus, they need a more efficient architecture to learn their correlations. Regarding the less spreading PM$_{10}$ values, our model is still better the Hybrid model in 12, 24 and 72 hours forecasting. From these experiments' results, we can conclude that the tight integration of GCN and RNN models in Spatiotemporal GCRNN architecture learns the spatiotemporal air pollution data better than separated layers in a GCN-based hybrid approach. Furthermore, the number of trainable parameters of the Hybrid model is 518K parameters, a \textbf{1.5x} larger our model.

\begin{table}
\renewcommand{\arraystretch}{1.2}
\caption{Statistical analysis of PM$_{2.5}$ and PM$_{10}$ air pollutants experimental data: Mean and Std. deviation (data is [0-1] normalized)}
\label{tab:stat_analysis} 
\centering
\begin{tabularx}{\linewidth}{| p{2cm}|p{1cm}p{1cm}|p{1.2cm}p{1.2cm} |}
\hline & \multicolumn{2}{c|}{PM$_{2.5}$} & \multicolumn{2}{c|}{PM$_{10}$} \\
\hline Statistical measurement & Mean & Std. deviation & Mean & Std. deviation  \\
\hline By time & 0.136927 & 0.091587 & 0.040974 & 0.031563 \\
\hline By locality & 0.136927 & 0.038279 & 0.040974 & 0.010903 \\
\hline
\end{tabularx}
\end{table}

\subsection{Ablation study the effects of different GCN configurations}
\label{gcn_configuration}
To evaluate the effects of different GCN configurations toward the forecasting results, we experiment with the following parameters: \textbf{$K$-order} to explore the impact of the graph convolutional filter's receptive fields; weighted adjacency matrix $W$ \textbf{threshold $\epsilon$} to vary the sparsity of $W$; and the \textbf{graph convolution definitions} (spectral or diffusion graph convolution).

\begin{figure}[!htb]
  \includegraphics[width=\linewidth]{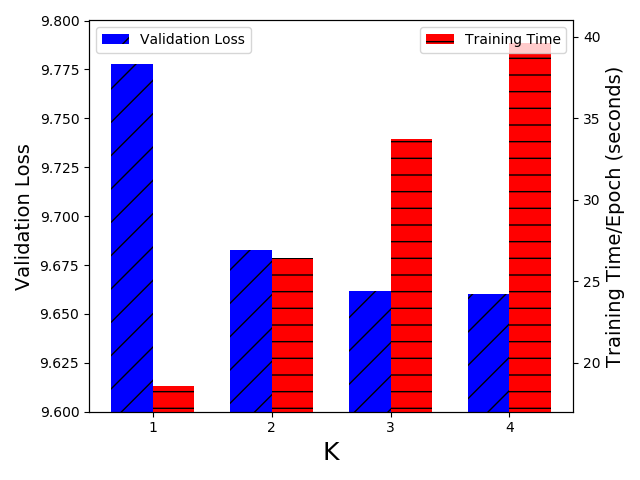}
  \caption{Effects of $K$-order values in Spatiotemporal GCRNN model. When K increases from 1 to 4, the Validation Loss decreases but the Training Time rises.}
  \label{fig:ex_compare_k}
\end{figure}

The \textbf{$K$-order} determines the number of nearest neighbors of a node that will be considered as the filter's receptive fields in graph convolutional operation. When $K$ increases, the size of receptive fields larger, and the graph can capture wider spatial dependency at the cost of increasing learning complexity and the training time. Fig. \ref{fig:ex_compare_k} shows the results with different $K$ on the validate set. We observe that with the increase of K, the validation loss first quickly falls ($K$ from 1 to 2), and then decreases slightly ($K \geq 3$), while the training time per epoch rises a lot. In this paper, we use \textbf{$K$ = 2} as the default $K$-order value to balance between the validation loss and the training time.

\begin{figure}[!htb]
  \includegraphics[width=\linewidth]{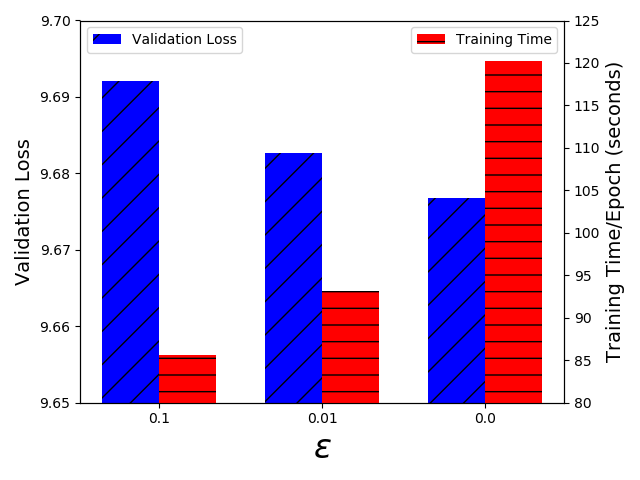}
  \caption{Effects of adjacency matrix threshold $\epsilon$ values in Spatiotemporal GCRNN model. When $\epsilon$ decreases from 0.1 to 0.0, the Validation Loss decreases but the Training Time also goes up.}
  \label{fig:ex_compare_epsilon}
\end{figure}

Regarding computing the weighted adjacency matrix $W$, we use a \textbf{threshold $\epsilon$} to control the sparsity of $W$. All weights less than $\epsilon$ will be set to zero. A weighted edge equals to zero also means there is no relationship between two nodes of that edge. In graph based problems, $W$ can be built using a thresholded Gaussian kernel \cite{Shuman2013}. $W_{ij} = exp(-\frac{dist(v_i, v_j)^2}{\sigma^2})$ if $dist(v_i, v_j) \leq \epsilon$, otherwise $0$. $dist(v_i, v_j)$ denotes the distance between 2 stations in station-level or 2 grid-cells in grid-level. $\sigma$ is the standard deviation of distances and $\epsilon$ is the threshold. In this paper, we experiment with three decreasing values of $\epsilon$: 0.1, 0.01, and 0.0 (no threshold). From the experiments' results in Fig. \ref{fig:ex_compare_epsilon}, we can observe that when $\epsilon$ decreases from 0.1 to 0.0, the validation loss decreases, but the training time for one epoch also goes up. Hence, we use $\mathbf{\epsilon = 0.01}$ as the default threshold in building the adjacency matrix for the graph since it has the trade-off between the validation errors and the training time.

\begin{figure}[!htb]
  \includegraphics[width=\linewidth]{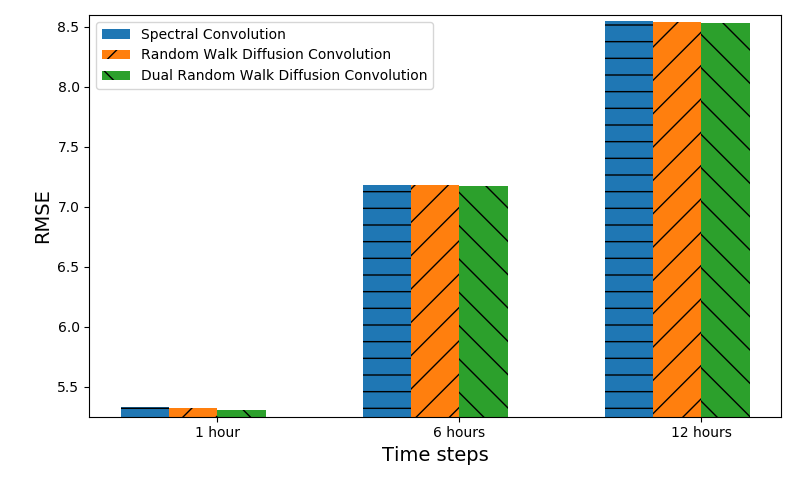}
  \caption{Performance comparisons for different graph convolution operators of PM$_{2.5}$ air pollution forecasting in 1, 6, and 12 hours ahead. Dual Random Walk Diffusion Convolution produces slightly better performance among two others.}
  \label{fig:ex_compare_conv}
\end{figure}

For the last ablation study, we explore the effects of \textbf{different graph convolution operators} such as \textit{spectral convolution} and \textit{diffusion convolution}. The spectral convolution operation is implemented as in equation \ref{eqn:spectral_conv} while the diffusion convolution is performed following equation \ref{eqn:diff_conv}. From \cite{DCRNN}, the diffusion convolution implementation includes two cases, the first is a \textit{random walk diffusion}, and the second is a \textit{dual random walk diffusion} with the diagonal matrix $D$ in equation \ref{eqn:diff_conv} is split into 2 matrices: an out-degree diagonal matrix and an in-degree diagonal matrix. Fig. \ref{fig:ex_compare_conv} shows RMSE values for these studying graph convolution operations with three time steps 1, 6, and 12 hours in PM$_{2.5}$ forecasting. As a results, the \textbf{dual random walk diffusion convolution} achieves a slightly better performance than other convolutional implementations.

\section{Conclusion}
\label{conclusions}
In this paper, we introduce a spatiotemporal GCN-based model for the citywide air pollution forecasting. The new proposal model named \textbf{Spatiotemporal Graph Convolutional Recurrent Neural Network} is a deep integration of a GCN model for spatial features extraction and an RNN model for temporal features learning. The Spatiotemporal GCRNN model has the size \textbf{55x smaller} than the state-of-the-art ConvLSTM model for air pollution forecasting but produces better results. Through a series of empirical experiments, we prove that our Spatiotemporal GCRNN model has better performance than the ConvLSTM model in short-term air pollution prediction, and also achieves superior results in medium to long-term air pollution forecasting compare to a GCN-based hybrid model that separates GCN and LSTM in discrete layers.

Moreover, in this research, we expand the previous Seoul data from 3 years time period to a 5-year dataset (\textbf{75$\%$ bigger}). We hope that this large-scale dataset will become a valuable data source for researchers in air pollution as well as spatiotemporal based research. We will public the new dataset along with this paper.

In the future, we will enhance our research in the following directions: first, exploiting advanced graph convolution network models such as spatial-based graph convolution or graph attention networks to achieve better performance; second, due to the small size of the graph-based model, we can deploy the trained model to build real-time spatiotemporal forecasting for urban intelligent systems.


\ifCLASSOPTIONcompsoc
  \section*{Acknowledgments}
\else
  \section*{Acknowledgment}
\fi

This work was supported by the New Industry Promotion Program (1415166913, Development of Front/Side CameraSensor for Autonomous Vehicle) funded by the Ministry of Trade, Industry Energy (MOTIE, Korea).




\begin{thebibliography}{1}

\bibitem{BigComp}
Van-Duc Le, Tien-Cuong Bui, and Sang-Kyun Cha, Spatiotemporal Deep Learning Model for Citywide Air Pollution Interpolation and Prediction, IEEE International Conference on Big Data and Smart Computing (BigComp) (2020).

\bibitem{YuZheng}
Yu Zheng, Xiuwen  Yi, Ming  Li, Ruiyuan  Li, Zhangqing  Shan, Eric  Chang, and Tianrui  Li, Forecasting fine-grained air quality based on big data, In Proceedings of the 21th ACM SIGKDD International Conference on Knowledge Discovery and Data Mining (pp. 2267-2276) (2015).

\bibitem{BIC2018}
Van Duc Le and Sang Kyun Cha, Realtime Air Pollution Prediction based on Spatiotemporal Big Data, The International Conference on Big data, IoT, and Cloud Computing (2018).

\bibitem{Alex2018Deep}
Tien-Cuong Bui, Van-Duc Le, and Sang-Kyun Cha, A Deep Learning Approach for Forecasting Air Pollution in South Korea Using LSTM, The International Conference on Big data, IoT, and Cloud Computing (2018).

\bibitem{Alex2020Star}
Tien-Cuong Bui, Joonyoung Kim, Taewoo Kang, Donghyeon Lee, Junyoung Choi, Insoon Yang, Kyomin Jung, and Sang Kyun Cha, STAR: Spatio-Temporal Prediction of Air Quality Using A Multimodal Approach, arXiv preprint arXiv:2003.02632 (2020).

\bibitem{SpectralGraph}
Joan Bruna, Wojciech Zaremba, Arthur Szlam, and Yann LeCun, Spectral Networks and Locally Connected Networks on Graphs, International Conference on Learning Representations, CBLS (2014).

\bibitem{ChebNet}
Michaël Defferrard, Xavier Bresson, and Pierre Vandergheynst, Convolutional Neural Networks on Graphs with Fast Localized Spectral Filtering, 30th Conference on Neural Information Processing Systems, Barcelona, Spain (2016).

\bibitem{SimplifiedChebNet}
Thomas N. Kipf, and Max Welling, Semi-Supervised Classification with Graph Convolutional Networks, International Conference on Learning Representations (ICLR) (2017).

\bibitem{DCRNN}
Yaguang Li, Rose Yu, Cyrus Shahabi, and Yan Liu, Diffusion Convolutional Recurrent Neural Network: Data-Driven Traffic Forecasting, International Conference on Learning Representations (2018).

\bibitem{STGCN}
Bing Yu, Haoteng Yin, and Zhanxing Zhu, Spatio-Temporal Graph Convolutional Networks: A Deep Learning Framework for Traffic Forecasting, Proceedings of the Twenty-Seventh International Joint Conference on Artificial Intelligence (2018).

\bibitem{DySTGCNN}
Zulong Diao, Xin Wang, Dafang Zhang, Yingru Liu, Kun Xie, and Shaoyao He, Dynamic Spatial-Temporal Graph Convolutional Neural Networks for Traffic Forecasting, Association for the Advancement of Artificial Intelligence (AAAI) (2019).

\bibitem{LinY}
Yijun  Lin, Nikhit  Mago, Yu  Gao, Yaguang Li, Yao-Yi Chiang, C. Shahabi, and José Luis Ambite, Exploiting spatiotemporal patterns for accurate air quality forecasting using deep learning, In Proceedings of the 26th ACM SIGSPATIAL international conference on advances in geographic information systems (pp. 359-368) (2018).

\bibitem{Hybrid}
Yanlin Qia, Qi Lia, Hamed Karimiana, and Di Liu, A hybrid model for spatiotemporal forecasting of PM$_{2.5}$ based on graph convolutional neural network and long short-term memory, Science of the Total Environment 664 (2019).

\bibitem{KnowledgeGNN}
Shuo Wang, Yanran Li, Jiang Zhang, Qingye Meng, Lingwei Meng, and Fei Gao, PM$_{2.5}$-GNN: A Domain Knowledge Enhanced Graph Neural Network For PM$_{2.5}$ Forecasting, In Proceedings of KDD’20: Under Review (2020).

\bibitem{LSTM}
Sepp Hochreiter and Jürgen Schmidhuber, Long short-term memory, Neural Computing 9(8): 1735-1780 (1997).

\bibitem{GRU}
Kyunghyun Cho, Dzmitry Bahdanau, Fethi Bougares, Holger Schwenk, and Yoshua Bengio, Learning Phrase Representations using RNN Encoder–Decoderfor Statistical Machine Translation, Proceedings of the 2014 Conference on Empirical Methods in Natural Language Processing (EMNLP), 1724–1734 (2014).

\bibitem{Seq2Seq}
Ilya Sutskever, Oriol Vinyals, and Quoc V Le, Sequence to sequence learning with neural networks, Advances in Neural Information Processing Systems (NIPS), pp. 3104–3112 (2014).

\bibitem{Tensorflow}
Martın Abadi et al., TensorFlow:Large-Scale Machine Learning on Heterogeneous Distributed Systems, Preliminary White Paper, arXiv:1603.04467 (2015).

\bibitem{Shuman2013}
David I Shuman, Sunil K Narang, Pascal Frossard, Antonio Ortega, and Pierre Vandergheynst, The emerging field of signal processing on graphs: Extending high-dimensional data analysis to networks and other irregular domains, IEEE Signal Processing Magazine, 30(3):83–98 (2013).

\end{thebibliography}
%

\newpage
\begin{wrapfigure}{l}{25mm} 
    \includegraphics[width=1in,height=1.25in,clip,keepaspectratio]{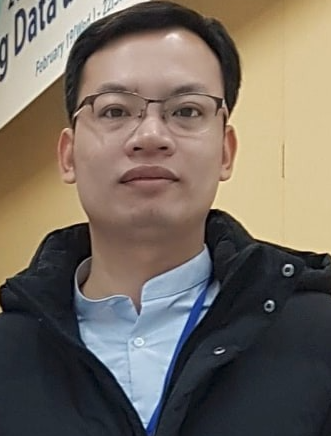}
  \end{wrapfigure}\par
  \textbf{Van-Duc Le} is a Ph.D. candidate at School of Electrical and Computer Engineering, Seoul National University, Seoul, South Korea. His research interests include Spatiotemporal Deep Learning, Ambient AI, and Machine Learning Life-Cycle Management. He received his MS from Seoul National University, Korea in 2019.\par

\begin{wrapfigure}{l}{25mm} 
    \includegraphics[width=1in,height=1.25in,clip,keepaspectratio]{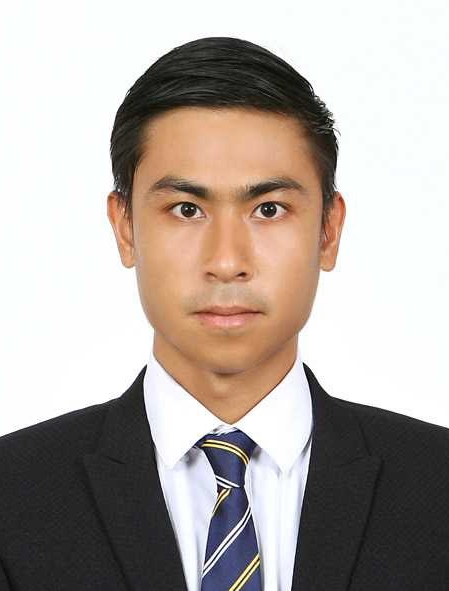}
  \end{wrapfigure}\par
  \textbf{Tien-Cuong Bui} is a Ph.D. candidate at School of Electrical and Computer Engineering, Seoul National University, Seoul, South Korea. His research interests include Data Mining, Natural Language Processing, Graph Mining, and Intelligent Infrastructure. He received his MS from Seoul National University, Korea in 2019.\par

\begin{wrapfigure}{l}{25mm} 
    \includegraphics[width=1in,height=1.25in,clip,keepaspectratio]{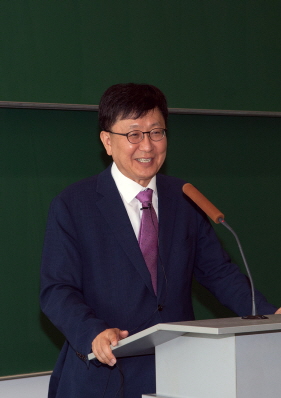}
  \end{wrapfigure}\par
  \textbf{Professor Sang-Kyun Cha} has been the founding dean of the Graduate School of Data Science of Seoul National University (SNU) since February, 2020. He led the effort of establishing this new graduate school to help transform Korea’s leading higher education institution in the age of data-driven innovation and AI since he took the role of founding director of SNU Big Data Institute in April, 2014. He received his BS and MS from Seoul National University and his Ph.D. from Stanford University.\par

\end{document}